\documentclass[letterpaper, 10 pt, conference]{ieeeconf}  

\usepackage{amsmath, mathtools, amssymb, bm}
\usepackage{tikz}
\usetikzlibrary{calc, angles, quotes}
\usepackage{pgfplots}
\pgfplotsset{compat=newest}
\usepgfplotslibrary{groupplots}
\usepackage{pgfplotstable}
\usepackage{subcaption}
\usepackage[labelfont=bf, font={footnotesize}]{caption}
\usepackage{xcolor}
\usepackage{siunitx}
\usepackage{lipsum}
\usepackage{multirow}
\usepackage[hidelinks]{hyperref}
\usepackage[capitalize]{cleveref}
\usepackage[english]{babel}

\newcommand{\todo}[1]{{\color{red}{TODO: #1}}}

\newcommand{\eqdef}{=\vcentcolon}

\newcommand{\tp}{^\mathsf{T}}
\newcommand{\bmd}[1]{\bm{\dot{#1}}}
\DeclareMathOperator*{\argmax}{arg\,max}

\newcommand{\los}{\bm{l}}

\newlength{\fwidth}
\newlength{\fheight}

\IEEEoverridecommandlockouts                              

\title{\LARGE \bf
Flying through Moving Gates without Full State Estimation
}

\author{Ralf R\"omer$^{1}$, Tim Emmert$^{1}$, and Angela P. Schoellig$^{1}$
\thanks{$^{1}$Technical University of Munich, Germany; TUM School of Computation, Information and Technology, Department of Computer Engineering, Learning Systems and Robotics Lab~(\href{www.learnsyslab.org}{learnsyslab.org}); Munich Institute of Robotics and Machine Intelligence~(MIRMI).}
\thanks{Corresponding Email: {\tt\footnotesize \{ralf.roemer; tim.emmert\}@tum.de}.}
\thanks{This work has been supported by the \href{https://www.robotics-institute-germany.de/}{Robotics Institute Germany}, funded by BMBF grant 16ME0997K.}
\thanks{R. R. gratefully acknowledges the support of the research group ConVeY funded by the German Research Foundation under grant GRK 2428.}
}

\begin{document}

\maketitle
\thispagestyle{empty}
\pagestyle{empty}

\begin{abstract}
Autonomous drone racing requires powerful perception, planning, and control and has become a benchmark and test field for autonomous, agile flight.
Existing work usually assumes static race tracks with known maps, which enables offline planning of time-optimal trajectories, performing localization to the gates to reduce the drift in visual-inertial odometry (VIO) for state estimation or training learning-based methods for the particular race track and operating environment.
In contrast, many real-world tasks like disaster response or delivery need to be performed in unknown and dynamic environments.
To make drone racing more robust against unseen environments and moving gates,
we propose a control algorithm that operates without a race track map or VIO, relying solely on monocular measurements of the line of sight to the gates.
For this purpose, we adopt the law of proportional navigation~(PN) to accurately fly through the gates despite gate motions or wind. 
We formulate the PN-informed vision-based control problem for drone racing as a constrained optimization problem and derive a closed-form optimal solution.
Through simulations and real-world experiments, we demonstrate that our algorithm can navigate through moving gates at high speeds while being robust to different gate movements, model errors, wind, and delays.

\end{abstract}

\section{Introduction}
Autonomous drone racing has become popular in the research community due to the availability of fast and agile quadrotors~\cite{foehn2022agilicious, hanover_autonomous_2024}. Since drone racing requires high-performance algorithms for perception, state estimation, planning, and control~\cite{tal_accurate_2020, kaufmann_champion-level_2023, qin_time-optimal_2024}, it is considered a challenging benchmark and a driver for progress in autonomous flight~\cite{hanover_autonomous_2024}.
The field has made tremendous progress recently, with maximum speeds evolving from~$2\,\si{m}/\si{s}$ in 2018~\cite{moon2019challenges} to more than~$25\,{\si{m}}/{\si{s}}$ in 2023, even matching human pilots~\cite{kaufmann_champion-level_2023}.

Prior work has almost exclusively considered the race track, i.e., the poses and sizes of the gates, to be known and static.
This has allowed for planning and tracking of time-optimal paths and trajectories~\cite{qin_time-optimal_2024, li_visual_2020, foehn2022alphapilot} or training learning-based state-estimation and control algorithms for the particular track~\cite{kaufmann_champion-level_2023, kaufmann2018deep}.
Moving gates are rarely addressed, with only one notable exception~\cite{kaufmann2018deep}. 
Even then, the 
perception/planning module cannot generalize to substantial changes in the race track or operating environment.
Onboard state estimation is commonly addressed via visual-inertial odometry~(VIO)~\cite{scaramuzza_visual-inertial_2019}. 
As VIO tends to accumulate large drift in situations involving motion blur, low-texture environments, and high dynamic range~\cite{foehn2022alphapilot}, VIO algorithms for high-speed drone racing use a race track map and perform localization relative to the gates to reduce drift~\cite{foehn2022alphapilot, kaufmann2018deep}.


The common assumptions of static and known operating environments 
differ from many real-world applications like search and rescue~\cite{schedl2021autonomous}, inspection of large structures, and delivery. In such scenarios, an accurate map is often unavailable, and drones can often only perceive their environment with monocular cameras.
Also, operating environments are often dynamic and subject to wind, requiring vision-based algorithms that can react to strong environmental changes and disturbances. 
The gaps between indoor drone racing on known and static tracks and applications in unknown and dynamic environments limit the applicability of solutions developed for drone racing on such real-world tasks.

Therefore, our goal in this work is to develop a drone racing algorithm that does not require a race track map and is robust to changes in the environment, particularly moving gates.
Specifically, we do not rely on VIO
to provide position and velocity information.
As a result, the relative position and velocity of the gate with respect to the quadrotor are unknown in our setup.
The only information about the gates we assume to be available to the control algorithm is an estimate of the gate's bounding box in the camera frame of a monocular camera mounted on the quadrotor, which can be obtained using standard detection algorithms~\cite{pham_2021_gatenet, pham_2022_pencilnet}. 

\input{figures/trajectory_planar2}
To fly through a moving gate with a quadrotor, the two trajectories must closely meet at a certain point in time.
We ensure this by 
using the concept of proportional navigation~(PN)~\cite{yanushevsky_lyapunov_2005}, which was originally developed as a missile guidance law for intercepting moving objects~\cite{yanushevsky2018modern}. 
The PN law cannot be directly applied to quadrotors 
due to several differences between missile and quadrotor control. Also, the PN law requires constant and known relative velocity, which does not match our objective of drone racing (i.e., minimizing total time) without velocity information. 
To address these gaps, we propose a novel PN-informed quadrotor control algorithm that 
is the first to pass moving gates using only monocular line-of-sight~(LOS) and IMU measurements. 
We visualize an example of flying through a fast-moving gate in~\cref{fig:trajectory_planar}.
In summary, our main contributions are:
\begin{itemize}
    \item We adapt the PN law for an unknown relative velocity and introduce the PN frame to derive a state-space model for PN-informed quadrotor control.
    \item We formulate PN-informed drone racing without full state estimation and with moving gates 
    as a constrained optimization problem and derive a closed-form solution.
    \item We present a Bayesian optimization~(BO) based hyperparameter optimization strategy to achieve robust performance for different objectives with our approach.
    \item We demonstrate through simulation and hardware experiments that our algorithm can pass moving gates based solely on LOS and IMU measurements 
    while being robust to different gate motions, 
    wind, and delays.
\end{itemize}

\section{Related Work}
\subsection{Autonomous Drone Racing}
Many drone racing approaches track a pre-planned path or trajectory that avoids obstacles and satisfies physical constraints~\cite{qin_time-optimal_2024, romero_model_2022, foehn_fast_2017, zhou_robust_2020, foehn_time-optimal_2021}. 
Common approaches for attitude control and trajectory tracking include nonlinear techniques~\cite{brescianini_tilt-prioritized_2020}, differential-flatness-based control~\cite{tal_accurate_2020} and model predictive control~(MPC)~\cite{romero_model_2022, sun_comparative_2024}.
Perception-aware planning and control approaches~\cite{greeff_perception-aware_2020, falanga_aggressive_2017, penin_vision-based_2017} can also consider the quality of the state estimates.
Planning-based methods require an accurate state estimate through motion capture or VIO~\cite{scaramuzza_visual-inertial_2019}. 
Without reliable estimation of the pose and velocity relative to the gate~\cite{foehn2022alphapilot, kaufmann2018deep}, which requires static gates at known locations, learning-based methods that skip the planning stage can be used~\cite{hanover_autonomous_2024, kaufmann_champion-level_2023}.
In~\cite{kaufmann2018deep}, a vision-based imitation learning policy is trained for trajectory tracking in a static environment, and the method can generalize to some gate motion after training.
Learning-based MPC with motion capture has been used to fly through a fast-moving gate~\cite{song2022policy}.
Swift~\cite{kaufmann_champion-level_2023}
uses a reinforcement learning~(RL)-based controller~\cite{schulman_proximal_2017} to generate rate commands and VIO~\cite{geneva_openvins_2020} for local position estimation on the track. 
The RL controller is trained for the particular race track, and the VIO algorithm requires a race track map and certain environmental conditions, which makes generalizing to changing or unseen operating environments challenging.
Recently, \cite{geles2024demonstrating} demonstrated agile flight at medium speeds without explicit state estimation, using the segmented gate edges as an input for an RL policy.
Despite recent advances in learning-based drone racing~\cite{kaufmann_champion-level_2023, foehn2022alphapilot, geles2024demonstrating}, these approaches have been limited to known and static race tracks in controlled environments.

\subsection{Proportional Navigation}
Proportional navigation~(PN)~\cite{yanushevsky_lyapunov_2005, yanushevsky2018modern} is a guidance law for intercepting moving objects based on keeping the bearing angle between the velocity vectors constant. 
Several recent studies~\cite{layman_evaluation_2021, kumar_real-time_2022, bhattacharya_toward_2021} address the adaptation of PN to quadrotors,  
but they rely on an accurate estimation of the relative distance and velocity, either via GPS~\cite{layman_evaluation_2021, bhattacharya_toward_2021} or ultrasonic sensors~\cite{kumar_real-time_2022}.
Therefore, these approaches are unsuitable for drone racing with moving gates based solely on LOS measurements.

\section{Problem Setup}
We consider the control problem of racing through moving gates without full state estimation.
The gates are detected with a monocular camera mounted on the quadrotor, and a (potentially noisy) estimate of the next gate's bounding box in the camera frame obtained via standard object detection algorithms~\cite{pham_2021_gatenet, pham_2022_pencilnet} is available.
Since objects in the real world often have unknown dimensions, we 
do not assume knowledge of the gate sizes, which could otherwise be used to estimate the relative distance to the gate.
Also, we do not use VIO for state estimation due to its reliance on known gate locations in high-speed drone racing scenarios~\cite{foehn2022alphapilot, kaufmann2018deep}. 
Consequently, the relative position and velocity between the quadrotor and the gate are assumed to be unknown.
As the lack of position and velocity information and the gate motion render the drone racing problem very difficult, we make three assumptions: 
1) The gates' top speeds are lower than the quadrotor's maximum speed.
2) The gate opening initially points towards the quadrotor, and the gate only exhibits translational motion.
3) After the quadrotor has passed a gate, the next gate is within the camera's field of view. We can then break down the problem of completing the entire race track into sequentially flying through the next gate in the field of view.

%
%

\section{Methodology}

\subsection{Preliminaries}

\subsubsection{Proportional Navigation} \label{sec:pre_pn}
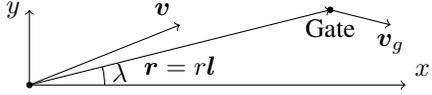
\begin{figure}
\centering
\begin{tikzpicture}

\draw[->] (0,0) -- (5,0) node[above right] {$x$};
\draw[->] (0,0) -- (0,1.2) node[left] {$y$};

\coordinate (O) at (0,0);
\filldraw (O) circle (1pt);

\coordinate (V) at (2,1);
\draw[->] (O) -- (V) node[above left] {$\bm{v}$};

\coordinate (Gate) at (4, 1.2); 
\filldraw (Gate) circle (1pt);
\node[below] at (Gate) {Gate};
\draw[->] (Gate) -- ++(0.8,-0.2) node[below] {$\bm{v}_g$};

\draw[->] (O) -- (Gate) node[midway, below] {$\bm{r} = r\los$};

\draw (1,0) arc (0:atan(0.3):1);
\node at (1.18,0.17) {$\lambda$};   

\end{tikzpicture}
\caption{Kinematics in the derivation of the PN law.}
\label{fig:pn_visualization}
\end{figure}

The PN law \cite{yanushevsky_lyapunov_2005, yanushevsky2018modern} 
was originally derived for use in interceptor missiles, but we consider a quadrotor and a moving gate in the following.
The PN law describes the planar kinematics in an inertial frame, as shown in~\cref{fig:pn_visualization}. We denote the quadrotor and gate velocity at time~$t$ by~$\bm{v}(t)$ and~$\bm{v}_\text{g}(t)$, but we omit the dependence on~$t$ for brevity.
The vector from the quadrotor to the gate center is denoted by~${\bm{r} = [x,\,y]\tp}$ and their distance by~${r=\|\bm{r}\|}$. 
We define the line-of-sight~(LOS) as~${\los=\frac{\bm{r}}{r}}$ and denote by~$\lambda$ the LOS angle between the~$x$-axis~$\bm{e}_x$ and~$\los$. 
The derivation of the PN law assumes~${x >> y}$ and a constant relative velocity~${v_\text{rel} = \dot{r}}$, i.e., $\ddot{r} = 0$. 
Then, we have ${\sin{(\lambda}) \approx \lambda = \frac{y}{r}}$, and the LOS angle evolves via~${\dot{\lambda} = \frac{1}{r}(\dot{y} - \lambda \dot{r})}$, ${\ddot{\lambda} = \frac{1}{r}(\ddot{y} - 2 \dot{\lambda} \dot{r} - \lambda \ddot{r})}$, 
%
%
where~${\ddot{y} = a_{\text{g},y}-a_{y}}$ is the difference between the accelerations~$a_{\text{g},y}$ and~$a_{y}$ of the gate and the quadrotor in~$y$-direction.
Defining the state as~${\bm{x} = [x_1,\,x_2]\tp = [\lambda,\,\dot{\lambda}]\tp}$, 
the control input as~${u=a_{y}}$ and the 
noise as~${w=a_{\text{g}, y}}$ and using~$\ddot{r}=0$ allows writing the LOS angle dynamics as
%
\begin{align}
    \label{eq:pre_engagement_kinematics_state_space}
    \begin{bmatrix}
        \dot{\lambda} \\ \ddot{\lambda}
    \end{bmatrix} =
    \begin{bmatrix}
        \dot{x}_1 \\ \dot{x}_2
    \end{bmatrix} = 
    \begin{bmatrix}
        \dot{x}_2 \\ \frac{1}{r} (w - u - 2 x_2 v_\text{rel})
    \end{bmatrix}.
\end{align}
%
According to the parallel navigation principle~\cite{yanushevsky2018modern}, the quadrotor will pass the gate center 
for a constant LOS angle~$\lambda$ and positive relative velocity,
i.e., if ${x_2 = 0}$ and ${\dot{r} > 0}$.
Assuming zero gate acceleration in~$y$-direction, i.e., ${a_{\text{g},y} = 0}$, a Lyapunov function ${V(\bm{x}) = \frac{1}{2}x_2^2}$ can be defined~\cite{khalil_nonlinear_2002}. Its time derivative satisfies~${\dot{V}(t) = x_2 \frac{1}{r} (- u - 2 x_2 v_\text{rel}) < 0}$
if 
\begin{equation}
    \label{eq:pre_pn_guidance_law}
    u(t) = k_\text{PN} v_\text{rel}(t) x_2(t), \quad k_\text{PN} > 2,
\end{equation}
for all~$t$, where~$k_\text{PN}$ is called the navigation constant. In summary, the PN law~\eqref{eq:pre_pn_guidance_law} commands an acceleration approximately normal to the LOS that guarantees stability of~\eqref{eq:pre_engagement_kinematics_state_space}. 

The PN-guidance approach can be generalized to 3D~\cite{yanushevsky2018modern}.
In this case, the LOS-rate is defined as~$\dot{\lambda} = ||\boldsymbol{\omega}_{\los}||$, where~$\bm{\omega}_{\los}$ is the angular velocity of~$\los$. The commanded acceleration is perpendicular to both~$\boldsymbol{\omega}_{\los}$ and~$\los$ and proportional to the norm of the relative velocity between the quadrotor and the gate.
We adopt the PN law for drone racing to ensure that a quadrotor flies precisely through and not past moving gates. 
\vspace{-0.4cm}
%
\subsubsection{Quadrotor Dynamics}
We define the quadrotor state as~${\bm{x} = \left(\bm{p}^\text{W}, \bm{q}_\text{B}^\text{W}, \bm{v}^\text{W}, \bm{\omega}^\text{B}\right)}$, where~$\bm{p}^\text{W}$ and~$\bm{v}^\text{W}$ are the position and velocity in the world frame~$\mathcal{O}_\text{W}$, $\bm{q}_\text{B}^\text{W}$ is the quaternion from the body frame~$\mathcal{O}_\text{B}$ to~$\mathcal{O}_\text{W}$, and~$\bm{\omega}^\text{B}$ are the body rates. 
The control input~${\bm{u} = \left(f_\text{th}, \bm{q}_\text{c}\right)}$ consists of the total thrust~$f_{\text{th}}$, acting in the positive~$z$-direction~$\bm{e}_z$ of~$\mathcal{O}_\text{B}$, and the desired orientation~$\bm{q}_\text{c}$.
We use the attitude controller~\cite{brescianini_nonlinear_2013} and approximate the PID rate controller~\cite{meier2015px4} as a first-order system with time constant~$\tau_\omega$. This yields the dynamics
\begin{align}
    \label{eq:meth_quadrotor_dynamics}
    \! \bmd{x} 
    = \bm{f}(\bm{x}, \bm{u})
    = \begin{bmatrix}
        \bmd{p}^\text{W} \\
        \bmd{q}_\text{B}^\text{W} \\
        \bmd{v}^\text{W} \\
        \bmd{\omega}^\text{B}
    \end{bmatrix} 
    = \begin{bmatrix}
        \bm{v}^\text{W} \\
        \bm{q}_\text{B}^\text{W} \cdot
        \begin{bmatrix}
            0 \\ \frac{1}{2}\bm{\omega}^\text{B}
        \end{bmatrix} \\
        \! \frac{1}{m} \bm{R}_\text{B}^\text{W} \! \left(f_\text{th}\bm{e}_z - \bm{f}_\text{aero}^\text{B}\right) + \bm{g}^\text{W} \! \! \\
        \frac{1}{\tau_{\omega}}\left(\bm{\omega}_c - \bm{\omega}^\text{B}\right)
    \end{bmatrix}\!\!, \!\!
\end{align}
where $m$~is the mass, ${\bm{R}_\text{B}^\text{W}}$~is the rotation matrix from~$\mathcal{O}_\text{B}$ to~$\mathcal{O}_\text{W}$, 
${\bm{g} = [g_x,\,g_y,\,g_z]\tp}$~is the gravity vector, 
$\bm{f}_\text{aero}$~are the aerodynamic forces, 
and~${\bm{\omega}_{\text{c}} = \frac{2}{\tau} \mathrm{sgn}(\bm{q}_{e, 0}) \mathbf{q}_{e, 1:3}}$ with the error quaternion~${\bm{q}_{\text{e}} = \left(\bm{q}_\text{B}^\text{W}\right)^{-1} \cdot \bm{q}_{\text{c}}}$~\cite{brescianini_nonlinear_2013}. 

\subsection{Monocular Line of Sight Estimation}
Adapting the PN law~\eqref{eq:pre_pn_guidance_law} for drone racing requires estimating the LOS~$\los$ and the LOS rate~$\dot{\lambda}$.
We use a monocular camera facing forward with zero camera pitch,
i.e., the optical axis~$\bm{o}_\text{cam}$ can be expressed in the body frame as~$\bm{o}_\text{cam}^\text{B} = \bm{e}_x$.
%
%
We detect the gate's bounding box in the camera frame~$\mathcal{O}_\text{C}$ at a frequency~$f_\text{est}$.
As these estimates are noisy due to motion blur, vibration, and tracking errors, we apply a low-pass filter. 
We calculate the LOS 
in~$\mathcal{O}_\text{C}$ from the bounding box center via the camera intrinsics and the LOS rate as
%
\begin{align}
    \dot{\lambda} = f_\text{est} \mathrm{arccos}\left(\los\tp \los_\text{prev} \right),
\end{align}
where~$\los_\text{prev}$ is the LOS estimate at the previous timestep.
We also compute the rotation axis of the LOS as~${\bm{k}_{\los} = \los \times \los_\text{prev}}$ to determine the direction in 3D space in which the acceleration given by the PN law must be applied, as explained below.

\subsection{Proportional Navigation without Velocity Measurements}
\begin{figure}
\centering
\begin{tikzpicture}[scale=1]
\draw[->] (0,0,0) -- (0,0,1) node[anchor=south] {$x$};
\draw[->] (0,0,0) -- (1,0,0) node[anchor=north east] {$y$};
\draw[->] (0,0,0) -- (0,1,0) node[anchor=north west] {$z$};

\coordinate (Gate) at (7,1,1);
\coordinate (Quadrotor) at (2,1,0);
\coordinate (Gravity) at (0, 2, 1);
\coordinate (quad_ez) at (0.34, 0.91, -0.23);   
\coordinate (o_cam) at (0.94, -0.24, 0.24);     
\coordinate (quad_ey) at (0.16, -0.30, -0.94);

\coordinate (o_cam0p5) at (0.47, -0.12, 0.12);
\coordinate (quad_ey0p5) at (0.08, -0.15, -0.47);
\coordinate (Propelleraxis12) at ($(o_cam0p5) + (quad_ey0p5)$);
\coordinate (Propelleraxis34) at ($(o_cam0p5) - (quad_ey0p5)$);
\coordinate (Propeller1) at ($(Quadrotor) - (Propelleraxis12)$);
\coordinate (Propeller2) at ($(Quadrotor) + (Propelleraxis12)$);
\coordinate (Propeller3) at ($(Quadrotor) - (Propelleraxis34)$);
\coordinate (Propeller4) at ($(Quadrotor) + (Propelleraxis34)$);

\definecolor{dronecolor}{rgb}{0, 0, 0} 

\draw[->] (Gravity) -- ++(0, -0.5, 0) node[left] {$\bm{g}$};

\draw[thick] (Gate) ellipse (0.3 and 0.9) node[above=0.9cm] {Gate};
\draw[thick] (Gate) ellipse (0.29 and 0.87);
\draw[thick] (Gate) ellipse (0.28 and 0.84);
\draw[thick] (Gate) ellipse (0.27 and 0.81);
\draw[thick] (Gate) ellipse (0.26 and 0.78);
\draw[thick] (Gate) ellipse (0.25 and 0.75);
\filldraw (Gate) circle (1pt);
\draw[->] (Gate) -- ++(0, 0, -1) node[right] {$\bm{v}_g$};

\fill[dronecolor] (Quadrotor) circle (0.03); 
\draw[-, thick] (Propeller1) -- (Propeller2);
\draw[-, thick] (Propeller3) -- (Propeller4);
\filldraw (Propeller1) circle (2pt);
\filldraw (Propeller2) circle (2pt);
\filldraw (Propeller3) circle (2pt);
\filldraw (Propeller4) circle (2pt);

\draw[->] (Quadrotor) -- ++(0.98, 0, 0.2) node[above left] {$\los$};    
\draw[-, dashed] (Quadrotor) -- (Gate); 
\draw[->] (Quadrotor) -- ++(0.10, 1.0, 0) node[left] {$\bm{k_l}$}; 
\draw[->] (Quadrotor) -- ++(-0.2, 0.02, 0.98) node[below] {$\bm{n}$};    
\draw[->] (Quadrotor) -- ++(o_cam) node[below] {$\bm{o}_\text{cam}$};
\draw[-, dashed] (Quadrotor) -- ++($2*(o_cam)$);

\coordinate (Endpoint1) at (3.96, 1, 0.39);      
\coordinate (Endpoint2) at (3.88, 0.52, 0.48);  

\pic [draw, -, "$\gamma$", angle radius = 1.5cm, angle eccentricity=1.2] {angle = Endpoint2--Quadrotor--Endpoint1};

\end{tikzpicture}
\caption{Proposed control approach for flying through moving gates solely based on measuring the line-of-sight~(LOS) angle~$\gamma$ between the optical axis of the camera~$\bm{o}_\text{cam}$ and the LOS~$\los$. 
To ensure that the gate is not missed, we command a PN-informed acceleration~$\bm{a}_n$ in the direction~$\bm{n}$ normal to~$\los$ and the LOS rotation axis~$\bm{k_l}$ via~\eqref{eq:meth_pn_guidance_law}. 
Moreover, we maximize the acceleration towards the gate, i.e., in the direction~$\bm{l}$, 
while enforcing an upper bound on~$\gamma$ to ensure that the gate always stays within the field of view.}
\label{fig:control_visualization}
\end{figure}
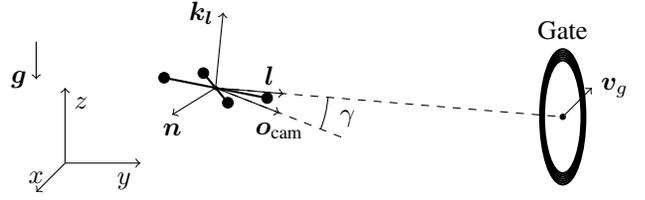

In the original PN law~\eqref{eq:pre_pn_guidance_law}, the relative velocity~$v_\text{rel}$ is assumed to be constant and known. 
Since the quadrotor in our setup can and should accelerate toward the gate to minimize time, and velocity estimates are unavailable, we need to modify~\eqref{eq:pre_pn_guidance_law}. 
Recall that the LOS angle dynamics~\eqref{eq:pre_engagement_kinematics_state_space} are stable if~$k_\text{PN} > 2$. 
Since the relative velocity~$v_\text{rel}$ is unknown in our setup, we define the lumped PN parameter~${k_v = k_\text{PN}v_\text{rel}}$, which needs to satisfy the condition~${k_v > 2v_\text{rel}}$ for all~$v_\text{rel}$.
We denote the maximum speed of the quadrotor as~$\bar{v}$ and consider the gate velocity to be upper bounded by~$\bar{v}_\text{g}$, which implies~${v_\text{rel} \leq \bar{v} + \bar{v}_\text{g} \eqdef \bar{v}_\text{rel}}$.
Then, the condition~${k_v > 2\bar{v}_\text{rel}}$ is sufficient for stability. 
The acceleration given by the PN law needs to be applied approximately normal to the LOS. 
Moreover, the derivation of the PN law
considers planar kinematics. 
To satisfy these conditions, we calculate the acceleration vector required for flying through the gate as
\begin{align}
    \label{eq:meth_pn_guidance_law}
    \bm{a}_n 
    = k_\text{PN} \bar{v}_\text{rel} \dot{\lambda} \bm{n}, \qquad \bm{n} 
    = \los \times \bm{k}_{\los},
\end{align}
where the acceleration is applied in the direction~$\bm{n}$ normal to the LOS~$\los$ and the LOS rotation axis~$\bm{k}_{\los}$, as visualized in~\cref{fig:control_visualization}.
In the following, we discuss how to generate this acceleration with a quadrotor at high speeds while meeting other requirements for vision-based drone racing.

\subsection{PN-Informed Quadrotor Control for Drone Racing} \label{meth_pn_for_quad}

\subsubsection{PN Frame}
%
%
The quadrotor PN law~\eqref{eq:meth_pn_guidance_law} commands an acceleration normal to the LOS. For drone racing, maximizing the speed towards the gate, i.e., along the LOS, is also desirable. 
For a unified formulation of these objectives, we define the PN frame~$\mathcal{O}_\text{PN}$ as
%
\begin{align}
    \label{eq:meth_rotation_pn_world}
    \bm{R}_\text{PN}^\text{W} = \begin{bmatrix}
        \los^{\text{W}} & \bm{e}_z \times \los^{\text{W}} & \los^{\text{W}} \times \left(\bm{e}_z \times \los^{\text{W}}\right)
    \end{bmatrix}.
\end{align}
As a result, we can express the LOS vector~$\los$ and the normal vector~$\bm{n}$ in~\eqref{eq:meth_pn_guidance_law} in~$\mathcal{O}_\text{PN}$ as
\begin{align}
    \los^{\text{PN}} = \bm{R}_\text{W}^{\text{PN}} \los^\text{W} = \bm{e}_x, \qquad
    \bm{n}^{\text{PN}} = \begin{bmatrix} 0 & n_y & n_z \end{bmatrix}\tp.
\end{align}
We assume that, as the quadrotor is approaching the gate, the aerodynamic forces act approximately parallel to the LOS, i.e., ${\bm{f}_\text{aero} \approx -f_\text{aero}\los}$, where~$f_\text{aero} = \|\bm{f}_\text{aero}\|$. Then, the quadrotor acceleration in the PN frame is given by
\begin{align}
    \label{eq:meth_quadrotor_acc_pn}
    \bm{a}^\text{PN} = \frac{1}{m} \left(\bm{R}_\text{B}^{\text{PN}}f_\text{th}\bm{e}_z - f_\text{aero}\bm{e}_x\right) + \bm{g}^\text{PN}.
\end{align}
This assumption is reasonable if the quadrotor is approximately on course to fly through the gate 
and if the gate velocity normal to the LOS is significantly smaller than the quadrotor velocity towards the gate. 

\subsubsection{Constrained Optimization Problem}
We aim to optimize 
the total thrust~$f_\text{th}$ and the commanded orientation~$\bm{q}_\text{c}$ with regard to the following conditions:
\begin{enumerate}
    \item The acceleration normal to the LOS must satisfy the quadrotor PN law~\eqref{eq:meth_pn_guidance_law} to fly through the moving gate.
    \item The quadrotor should achieve high speed towards the gate to minimize lap time.
    \item While flying towards the gate, the gate must always stay in the camera's field of view.
\end{enumerate}
We also consider an upper thrust limit~$\bar{f}_\text{th}$ and formulate the above requirements as a constrained optimization problem
\begin{subequations}
\label{eq:meth_optimization_problem_original}
\begin{align}
    \label{eq:meth_optimization_problem_original_cost}
    f_\text{th}^*, \bm{q}_\text{c}^* = \argmax_{f_\text{th}, \bm{q}_c} \quad & a_{\text{los}} \\
    \label{eq:meth_optimization_problem_original_acc}
    \text{s.t.} \qquad  & \bm{a}^\text{PN} = 
                            \begin{bmatrix}
                                a_\text{los} & n_y a_n & n_z a_n
                            \end{bmatrix}\tp \\
                        \label{eq:meth_optimization_problem_original_thrust}
                        & f_\text{th} \in \left[0, \bar{f}_\text{th}\right] \\
                        \label{eq:meth_optimization_problem_original_orientation}
                        & \bm{q}_\text{c} \in \mathcal{Q}_{\text{fov}},
\end{align}
\end{subequations}
where~$\bm{a}^\text{PN}$ is given by~\eqref{eq:meth_quadrotor_acc_pn}, ${a_n = \|\bm{a}_n\|}$ with~$\bm{a}_n$ given by~\eqref{eq:meth_pn_guidance_law}, and~$\mathcal{Q}_\text{fov}$ contains all orientations keeping the gate within the field of view.
To simplify this nonlinear optimization problem, we use the fact that accelerating towards the gate requires pitching, and applying the normal acceleration~\eqref{eq:meth_pn_guidance_law} requires rolling, both with respect to the PN frame~$\mathcal{O}_\text{PN}$. Consequently, we can decompose the rotation from the body frame~$\mathcal{O}_\text{B}$ to~$\mathcal{O}_\text{PN}$ into two rotations as
\begin{align}
    \label{eq:meth_body_pn_rotation}
    \bm{R}_\text{B}^\text{PN} = \bm{R}_y(\phi) \bm{R}_x(\theta),
\end{align}
where~$\phi$ and~$\theta$ are the pitch and roll angle, respectively. By 
inserting~\eqref{eq:meth_body_pn_rotation} into~\eqref{eq:meth_quadrotor_acc_pn}, we can rewrite~\eqref{eq:meth_optimization_problem_original} as
\begin{subequations}
\label{eq:meth_optimization_problem_simplified}
\begin{align}
    \label{eq:meth_optimization_problem_simplified_cost}
    \argmax_{f_\text{th}, \phi, \theta} \quad & f_\text{th} \sin{(\phi)} \cos{(\theta)} \\
                        \label{eq:meth_optimization_problem_simplified_cost_ay_az}
                        & \begin{bmatrix}
                            n_y a_n \\ n_z a_n
                        \end{bmatrix} = 
                        \begin{bmatrix}
                            -f_\text{th} \sin{(\theta)} \\
                            f_\text{th} \cos{(\phi)} \cos{(\theta)} + g_z^\text{PN}
                        \end{bmatrix} \\
                        & \eqref{eq:meth_optimization_problem_original_thrust}, \; \eqref{eq:meth_optimization_problem_original_orientation} \nonumber,
\end{align}
\end{subequations}
where we have used that~${g^{\text{PN}}_{y} = 0}$ due to the definition of~$\mathcal{O}^\text{PN}$ in~\eqref{eq:meth_rotation_pn_world} and that~$f_\text{aero}$ and $g_x^{\text{PN}}$ are independent of 
$f_\text{th}$, $\phi$ and $\theta$. Solving~\eqref{eq:meth_optimization_problem_simplified_cost_ay_az} for the roll and pitch angle yields
\begin{subequations}
\label{eq:meth_optimization_problem_roll_pitch}
\begin{align}
    \label{eq:meth_optimization_problem_roll}
    \theta(f_\text{th}) &= \arcsin{\left(- \frac{n_y a_n}{f_\text{th}}\right)} \\
    \label{eq:meth_optimization_problem_pitch}
    \phi(f_\text{th}) &= \arccos{\left(
    \frac{n_z a_n - g_z^{\text{PN}}}{\sqrt{f_\text{th}^2 - (n_y a_n)^2}}
    \right)},
\end{align}
\end{subequations}
which we can use to rewrite the objective~\eqref{eq:meth_optimization_problem_simplified_cost} as
\begin{align}
    \label{eq:meth_optimization_problem_cost_solution}
    f_\text{th} \sin{(\phi)} \cos{(\theta)} = \sqrt{f_\text{th}^2 - n_y^2 a_n^2 - (n_z a_n - g_z^\text{PN})^2}.
\end{align}
%
To keep the gate in the field of view, we impose an upper bound~${\bar{\gamma} \in (0, \gamma_\text{cam}]}$ on the angle~$\gamma$ between the LOS~$\los$ and the optical axis~$\bm{o}_\text{cam}$ of the camera, where~$\gamma_\text{cam}$ is the camera's angle of view. 
Increasing~$\bar{\gamma}$ increases the risk that due to the gate motion, model errors, or disturbances, the gate leaves the field of view at some point and can no longer be detected. 
It follows from
~${\bm{o}_\text{cam}^\text{B}=\bm{e}_x}$
and~\eqref{eq:meth_body_pn_rotation} that the optical axis in the PN frame~$\mathcal{O}_\text{PN}$ is given by
%
\begin{align}
    \bm{o}_\text{cam}^{\text{PN}} 
    = \bm{R}_\text{B}^{\text{PN}} \bm{o}_\text{cam}^{\text{B}} 
    = \bm{R}_y(\phi) \bm{R}_x(\theta) \bm{e}_x.
\end{align}
Hence, we can calculate~$\gamma$ 
in~$\mathcal{O}_\text{PN}$ via
\begin{align*}
    \cos{(\gamma)}  
    = \bm{l}\tp \bm{o}_\text{cam} 
    = \bm{e}_x\tp \bm{R}_y(\phi) \bm{R}_x(\theta) \bm{e}_x 
    = \cos{(\phi)} \cos{(\theta)}.
\end{align*}
By inserting~\eqref{eq:meth_optimization_problem_roll_pitch}, we can express~${\gamma < \bar{\gamma}}$ as
\begin{align}
    \label{eq:meth_optimization_problem_field_of_view_solution}
    \frac{n_z a_n - g_z^\text{PN}}{f_\text{th}} \geq \cos{(\bar{\gamma})}.
\end{align}
It follows from~\eqref{eq:meth_optimization_problem_roll_pitch}, \eqref{eq:meth_optimization_problem_cost_solution} and~\eqref{eq:meth_optimization_problem_field_of_view_solution} that the optimal control input solving the optimization problem~\eqref{eq:meth_optimization_problem_simplified} is given by
\begin{align*}
    f_\text{th}^* &= \min{\left(\bar{f}_\text{th}, \frac{n_z a_n - g_z^\text{PN}}{\cos{(\bar{\gamma})}}\right)}, \;\;
    \phi^* = \phi(f_\text{th}^*), \;\;
    \theta^* = \theta(f_\text{th}^*).
\end{align*}
We can see that increasing the upper bound~$\bar{\gamma}$ also makes the control towards the gate more aggressive.

\subsection{Bayesian Hyperparameter Optimization} \label{sec:bo}
The racing performance of our control algorithm is affected by two key hyperparameters: The navigation constant $k_\text{PN}$ and the upper bound~$\bar{\gamma}$ on the angle between the camera's optical axis and the LOS.
To reduce the tuning effort, we use Bayesian optimization~(BO)~\cite{brochu_tutorial_2010}, a common approach for optimizing black-box functions that are expensive to evaluate. 
We quantify the performance of a set of hyperparameters via two metrics: The time to reach the gate~$t_\text{gate}$ and the distance~$d_\text{center}$ to the gate center at the moment of flying through it.
The choice to minimize~$d_\text{center}$ is made for two reasons: First, passing the gate close to its center increases the robustness against colliding with the gate boundaries. 
Second, due to the lack of position information and the gate movements, we cannot track an optimal trajectory~\cite{qin_time-optimal_2024, romero_time-optimal_2022}, which would not necessarily go through the gate centers.
To obtain optimal hyperparameters with respect to the two metrics, we define a reward function
\begin{align}
    \label{eq:meth_bo_reward}
    R(k_\text{PN},\,\bar{\gamma}) = \max{\left(0, \ c_t / t_{\text{gate}} - c_d d_\text{center} + c\right)},
\end{align}
where~$c_t \geq 0$ and~$c_d \geq 0$ are weighting factors to trade off the two objectives, and~${c \geq 0}$ is an offset.
In each iteration of BO, we first fit a Gaussian process~(GP) model~\cite{rasmussen_gaussian_2006} to the available parameter-reward samples.
Then, we determine the next set of hyperparameters~$\bm{\theta}$ to evaluate by maximizing the upper confidence bound~\cite{brochu_tutorial_2010} of the fitted GP by utilizing its mean and variance. After a few iterations, we obtain optimized hyperparameters for our control algorithm.

\section{Evaluation}

\subsection{Simulation}
\subsubsection{Setup}
\begin{figure}[tb] 
    \centering
        \includegraphics[width=0.49\columnwidth, trim=0 5 0 70, clip]{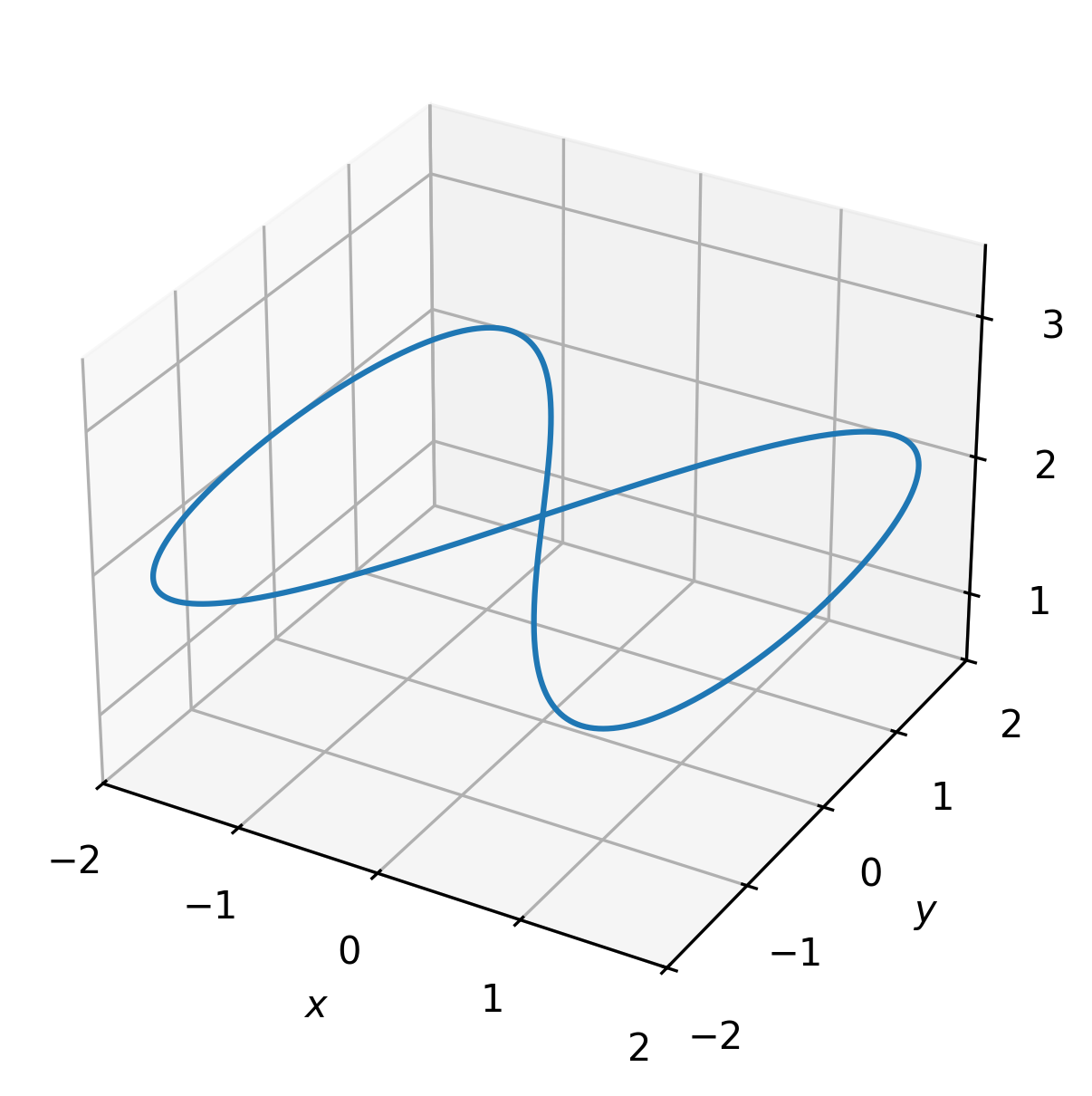}
        \includegraphics[width=0.49\columnwidth, trim=0 5 0 70, clip]{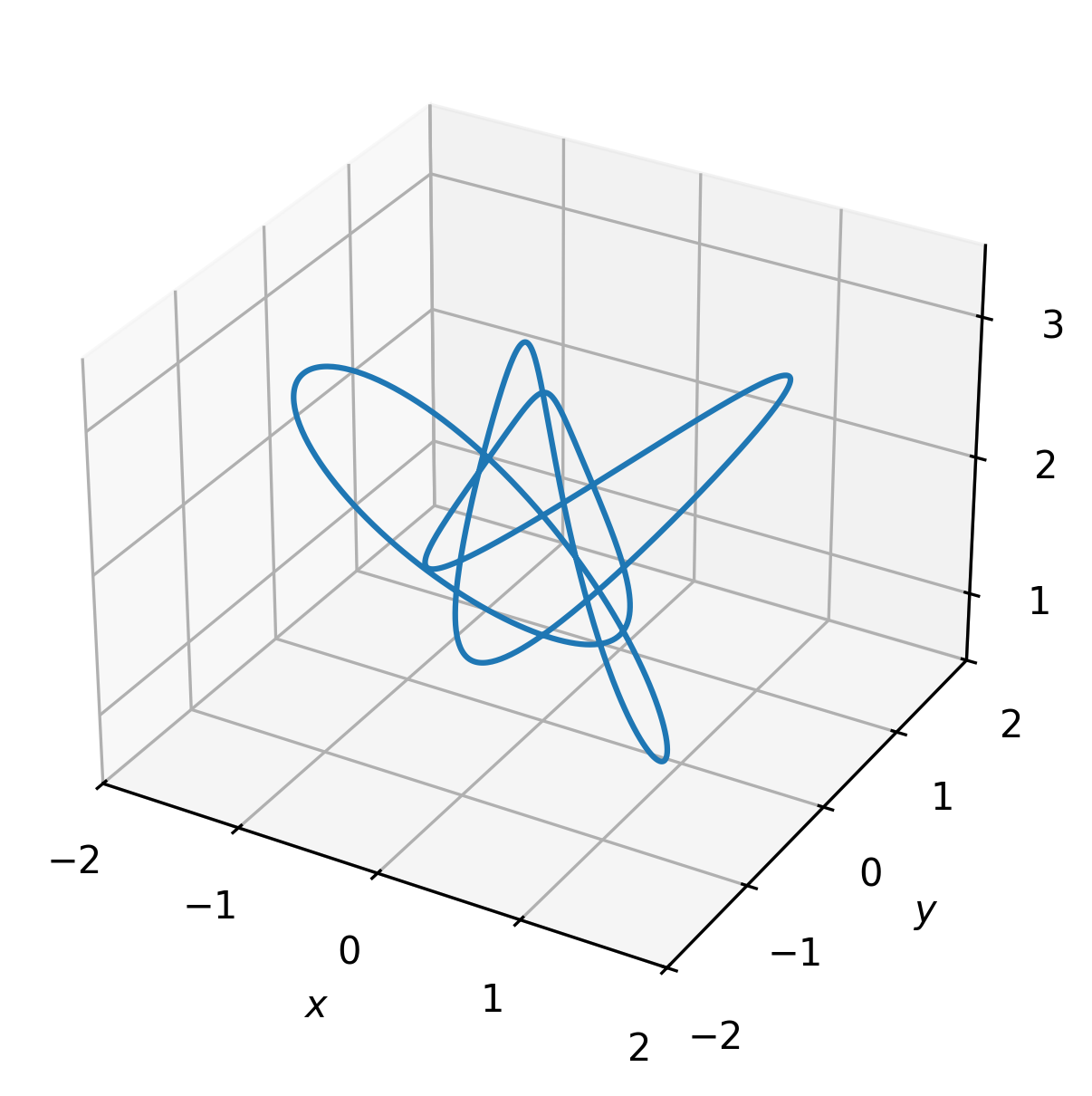}
    \caption{Planar~(2D) and knot~(3D) gate motions used in our experiments.}
    \vspace{-4pt}
    \label{fig:motion_overview}
\end{figure}

We utilize four parameterizations of the gate position over time:
Stationary~(0D), linear~(1D), planar~(2D), and knot~(3D). 
We use two different constant velocities for the linear motion, referred to as ``slow''~($2.5\,\si{m}/\si{s}$) and ``fast''~($5\,\si{m}/\si{s}$).
The planar and knot motions are defined by
\begin{align*}
    \bm{p}_\text{plan}(t) &= \bm{p}_0 + d_\text{plan}\begin{bmatrix}
        \mathrm{c}_{\omega t} & \frac{1}{2} \mathrm{c}_{2 \omega t} & 0
    \end{bmatrix}\tp, \\
    \bm{p}_\text{knot}(t) &= \bm{p}_0 + d_\text{knot}\begin{bmatrix}
        \mathrm{s}_{3 \omega t} \left(1 + \frac{1}{2}\mathrm{c}_{2 \omega t}\right) & \mathrm{s}_{3 \omega t} \mathrm{s}_{2 \omega t} & \mathrm{s}_{4 \omega t}
    \end{bmatrix}\tp,
\end{align*}
where~${\mathrm{c}_{\alpha} = \cos{(\alpha)}}$, ${\mathrm{s}_{\alpha} = \sin{(\alpha)}}$,
and shown in~\cref{fig:motion_overview}. Unless otherwise stated, we set~${d_\text{plan}=2\,\si{m}}$, ${d_\text{knot}=1\,\si{m}}$, resulting in a top speed and acceleration of~${3.1\,\si{m}/\si{s}}$ and~${4.0\,\si{m}/\si{s}^2}$ for the planar and~${1.2\,\si{m}/\si{s}}$ and~${0.9\,\si{m}/\si{s}^2}$ for the knot motion.

We combine the software-in-the-loop simulation of PX4~\cite{meier2015px4} and Gazebo to simulate the quadrotor dynamics, including rotors~\cite{koenig_design_2004, martin_true_2010}, low-level control, and aerodynamic forces, as well as our drone racing algorithm for the x500 quadrotor~\cite{meier2015px4} with a mass of~$2\,\si{kg}$.
The simulation provides simulated sensor measurements by applying noise to the ground truth acceleration, angular velocity, and air pressure for the IMU and barometer sensors.
We use the quadrotor's true position and attitude to project the gate's ground truth position into the camera frame based on its calibration. 
To evaluate robustness, imperfections in the vision system can be simulated by 
adding noise or 
introducing a delay to the bounding box measurements. 
We use an estimation and control frequency of~$30\,\si{Hz}$ in all experiments. The camera angle of view is~${\gamma_\text{cam} = 33.5^\circ}$. The gates are circular with a radius of~$0.5\,\si{m}$ to~$2\,\si{m}$.
Unless otherwise stated, we consider initial distances from the quadrotor to the gate between~$20\,\si{m}$~and~$30\,\si{m}$, set the initial velocity to~$10\,\si{m}/\si{s}$ and randomize the angle between the initial velocity vector and the LOS between~$-10\,\si{^\circ}$ and~$10\,\si{^\circ}$.
For the planar and knot gate motion, we randomize the initial LOS angle by initializing the gate position randomly on the gate trajectory.

\subsubsection{Bayesian Hyperparameter Optimization}

We optimize~$k_\text{PN}$ and~$\bar{\gamma}$ via BO~\cite{bo_toolbox_2014}, as described in~\cref{sec:bo}.
For the GP, we use a Matérn kernel~\cite{rasmussen_gaussian_2006} with smoothness parameter~$\nu = 2.5$.
The estimate of the maximum relative speed used in~\eqref{eq:meth_pn_guidance_law} is~${\bar{v}_\text{rel}=15\,\si{m}/\si{s}}$. We set the search spaces for the hyperparameters to~${\bar{\gamma} \in [5^\circ,\,30^\circ]}$ and~${k_\text{PN}\in[0.3,\,3.0]}$.
We consider two reward formulations.
The ``distance + time'' reward~$R_{d,t}$ uses~${c_t = 10}$, ${c_d = 3}$ and ${c = 0}$ in~\eqref{eq:meth_bo_reward}, aiming to strike a balance between robustness against collisions with the gate and minimizing time.
The ``distance'' reward~$R_d$ reward uses~${c_t = 0}$, ${c_d = 5}$, and ${c=3}$, corresponding to a conservative ``racing'' approach where the primary goal is to successfully fly through the gates. 
We perform $25$~iterations for BO, each consisting of~$2$ runs per gate motion (excluding the stationary gate as it is the easiest to fly through).
The optimized hyperparameter values are~${k_\text{PN}=2.10}$ and~${\bar{\gamma}=21.05^\circ}$ for the reward~$R_{d,t}$ and~${k_\text{PN}=1.11}$ and~${\bar{\gamma}=8.91^\circ}$ for the reward~$R_d$.
This shows that the combined reward~$R_{d,t}$ results in a more aggressive control behavior and allows having the gate closer to the boundary of the field of view. 

\subsubsection{Results and Discussion}
\setlength{\tabcolsep}{5pt}
\begin{table}[tb!]
\centering
\begin{tabular}{ccccc}
\hline
\multirow{ 2}{*}{Gate motion} & \multicolumn{2}{c}{Success rate} & \multicolumn{2}{c}{Time~$\left[\si{s}\right]$} \\
& Rew. $R_{d,t}$ & Rew. $R_d$ & Rew. $R_{d,t}$ & Rew. $R_d$ \\
\hline
Stationary & $1.00\;(1.00)$  & $1.00\;(1.00)$ & $4.11$ & $5.28$ \\
Linear slow & $1.00\;(1.00)$ & $1.00\;(1.00)$ & $4.03$ & $5.35$ \\
Linear fast & $0.67\;(0.75)$ & $0.92\;(1.00)$ & $5.41$ & $5.80$ \\
Planar & $1.00\;(1.00)$ & $0.50\;(1.00)$ & $4.01$ & $5.76$ \\
Knot & $1.00\;(1.00)$ & $1.00\;(1.00)$ & $4.03$ & $5.41$ \\
\hline
Total & $\bm{0.93}\;(0.95)$ & $0.88\;(\bm{1.00})$ & $\bm{4.32}$ & $5.52$ \\
\hline
\end{tabular}
\caption{Performance of our algorithm for racing through moving gates for two different formulations of the reward~\eqref{eq:meth_bo_reward}. The success rates of flying through the gates are reported for a gate radius of~$1\,\si{m}$ and~$2\,\si{m}$ (in brackets).}
\label{tab:racing_performance}
\vspace{-1pt}
\end{table}

\setlength{\tabcolsep}{6pt}

%
We simulate~$60$ runs with different initial conditions for both reward formulations, evenly spread across all gate motions.
The success rate of flying through the gates and the time to reach the gates~$t_\text{gate}$ are provided in~\cref{tab:racing_performance}. 
Our approach achieves a high success rate of at least~$88\%$, and the same hyperparameters perform well for different gate motions. 
The linear fast and planar gate motions are the most challenging, likely because their high speeds violate some of the assumptions in the derivation of the PN law in~\cref{sec:pre_pn}. 
Although~${k_\text{PN} \leq 2}$ for the reward~$R_d$, high performance is achieved due to overestimating the relative velocity. 
The combined reward~$R_{d,t}$ leads to shorter times to pass the gates and higher top speeds ($14.6\,\si{m}/\si{s}$ vs. $11.0\,\si{m}/\si{s}$), and we use it in the following.


%
%
\pgfplotsset{compat=1.8}
\begin{figure}[tb!]
\centering
\setlength{\fwidth}{5.6cm}
\setlength{\fheight}{1.6cm}
\definecolor{mycolor1}{RGB}{31,119,180}
\definecolor{mycolor2}{RGB}{214,39,40}
\definecolor{mycolor3}{RGB}{44,160,44}
\definecolor{mycolor4}{RGB}{255,127,14}
\definecolor{mycolor5}{RGB}{148,103,189}
\begin{tikzpicture}

\begin{axis}[
width=\fwidth,
height=\fheight,
ticklabel style = {font=\footnotesize},
label style = {font=\footnotesize},
title style = {font=\footnotesize},
scale only axis,
xmin=-0.01,
xmax=0.31,
xlabel={Time delay~$T_\text{bb}$ $\left[\si{s}\right]$},
xlabel style={yshift=2pt},
ymin=0,
ymax=1.1,
ylabel={Success rate},
ylabel style={align=center, yshift=-2pt},
ytick={0,0.2,0.4,0.6,0.8,1},
legend style={at={(0,0)}, anchor=south west, legend cell align=left, legend columns=3, align=left,font=\footnotesize,legend image post style={scale=1.0}},
]

\addplot[mycolor1, thick, mark=+, only marks]
 table[row sep=crcr]{%
    0 0.63 \\
    0.1 0.44 \\
    0.2 0.48 \\
    0.3 0.28 \\
}; 
\addlegendentry{$r_\text{gate} = 0.5\,\si{m}$}

\addplot[mycolor2, thick, mark=x, only marks]
 table[row sep=crcr]{%
    0 0.96 \\
    0.1 0.93 \\
    0.2 0.63 \\
    0.3 0.52 \\
}; 
\addlegendentry{$r_\text{gate} = 1\,\si{m}$}

\addplot[mycolor3, thick, mark=o, only marks]
 table[row sep=crcr]{%
    0 1 \\
    0.1 1 \\
    0.2 1 \\
    0.3 0.93 \\
}; 
\addlegendentry{$r_\text{gate} = 2\,\si{m}$}

\end{axis}

\end{tikzpicture}%

\caption{Impact of delays in the vision-based LOS estimation (e.g., due to camera latencies, inference times) on the success rate of passing the gates.}
\label{fig:robustness_latency}
\vspace{-5pt}
\end{figure}
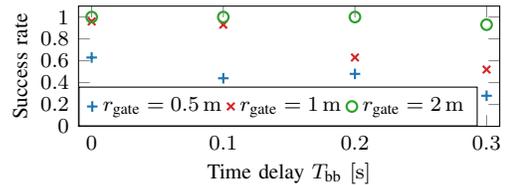
We also evaluate the robustness of our algorithm against 
time delays in the perception, using the stationary, linear fast, and planar gate motions.
For this, we consider delays of up to~$300\,\si{ms}$ in the LOS estimation.
As shown in~\cref{fig:robustness_latency}, even a large delay of~$100\,\si{ms}$ has little effect on performance for gate radii of~$1\,\si{m}$ and~$2\,\si{m}$.

One of the key properties of our control algorithm is that it does not require knowledge of the relative speed between the quadrotor and the gate and instead uses a fixed overestimation~$\bar{v}_\text{rel}$. 
We compare this approach to the ideal case when the relative speed is known.
To obtain very different relative speeds, we consider initial distances to the gate of~$30\,\si{m}$ and~$150\,\si{m}$, leading to top speeds of around~$15\,\si{m}/\si{s}$ and~$25\,\si{m}/\si{s}$. We use the knot gate motion and set~${d_\text{knot}=5}$ to match the increased initial distance.
The results for different estimates~$\bar{v}_\text{rel}$ are provided in~\cref{tab:relative_velocity_estimation}. We observe that performance is poor if the maximum relative speed is vastly over- or underestimated. However, if the estimate is reasonable, our algorithm performs comparably to the idealized case, highlighting the robustness of our approach to different gate motions.  
We observe that using the true relative speed performs worse for the higher top speed, likely because that approach violates the assumption of a constant factor~$k_\text{PN}v_\text{rel}$ in the original PN law~\eqref{eq:pre_pn_guidance_law}.
\setlength{\tabcolsep}{6pt}

\begin{table}[tb!]
\centering
\begin{tabular}{ccc}
\hline
Top speed [$\si{m}/\si{s}$] & Knowledge about~$v_\text{rel}$ & $d_\text{center}$ [$\si{m}$] \\
\hline
\multirow{4}{*}{$15$}   & Estimate: $\bar{v}_\text{rel} = 30\,\si{m}/\si{s}$ & $4.72 \pm 1.03$ \\
                        & Estimate: $\bar{v}_\text{rel} = 20\,\si{m}/\si{s}$ & $1.01 \pm 0.22$ \\
                        & Estimate: $\bar{v}_\text{rel} = 10\,\si{m}/\si{s}$ & $1.02 \pm 0.34$ \\
                        & Perfect knowledge & $\bm{0.42\pm 0.19}$ \\
\hline
\multirow{4}{*}{$25$}   & Estimate: $\bar{v}_\text{rel} = 30\,\si{m}/\si{s}$ & $\bm{0.65 \pm 0.05}$ \\
                        & Estimate: $\bar{v}_\text{rel} = 20\,\si{m}/\si{s}$ & $0.70 \pm 0.35$ \\
                        & Estimate: $\bar{v}_\text{rel} = 10\,\si{m}/\si{s}$ & $4.84 \pm 0.45$ \\
                        & Perfect knowledge & $1.19\pm 0.43$ \\
\hline
\end{tabular}
\vspace{-1pt}
\caption{Comparing our approach to overestimate the relative velocity~$v_\text{rel}$ to be~$\bar{v}_\text{rel}$ with the idealized case that~$v_\text{rel}$ is known at all times.}
\label{tab:relative_velocity_estimation}
\end{table}

\setlength{\tabcolsep}{6pt}

The only baseline conducting experiments comparable to ours is~\cite{bhattacharya_toward_2021}, which directly commands the acceleration from the PN law to the PX4 acceleration controller. 
In contrast to our approach, \cite{bhattacharya_toward_2021} assumes a velocity estimate to be available via GPS and controls the speed to a relatively low, constant value of~$5\,\si{m}/\si{s}$.
For a fair comparison, we constrain the pitch angle by setting~${\bar{\gamma} = 3.5^\circ}$
to achieve a similar top speed and use the same gate radius of~$0.5\,\si{m}$ and gate motion parameterizations as~\cite{bhattacharya_toward_2021}, with gates moving at~$1.25$ to~$5\:\si{m}/\si{s}$. 
The success rates of flying through the gates given in~\cref{tab:baseline_comparison} show that our PN-informed controller clearly outperforms the approach~\cite{bhattacharya_toward_2021} 
although we assume strictly fewer parts of the state estimate to be available.
\setlength{\tabcolsep}{4pt}

\begin{table}[tb!]
    \centering
    \begin{tabular}{c c c}
        \hline
        Gate motion & PN-based accel. contr.~\cite{bhattacharya_toward_2021} & PN-informed contr. (ours) \\
        \hline
        Linear & $0.35$ & $\bm{0.69}$ \\
        Planar & $0.41$ & $\bm{0.56}$ \\
        \hline
    \end{tabular}
    \vspace{-1pt}
    \caption{Comparison against~\cite{bhattacharya_toward_2021}, which assumes velocity information to be available, in terms of the success rate of passing a~$0.5\,\si{m}$ radius gate.} 
    \label{tab:baseline_comparison}
\end{table}

\setlength{\tabcolsep}{6pt}

\subsection{Hardware Experiments}
\subsubsection{Setup}
We conduct real-world experiments in an uncontrolled outdoor environment with a quadrotor with~$2\,\si{kg}$ mass and~$25\,\si{m}/\si{s}$ maximum speed.
We implement the LOS calculation and the control algorithm in ROS2~\cite{macenski2022robot} and use a UXRCE-DDS bridge for communication with PX4.

\subsubsection{Simulated Gate Visuals}
\begin{table}[tb!]
\centering
\begin{tabular}{cccc}
\hline
Runs & Distance to gate center [$\si{m}$] & Top speed [$\si{m}/\si{s}$] & Time [$\si{s}$] \\
\hline
$1$ & $0.39$ & $13.36$ & $4.21$ \\
$2$ & $0.30$ & $12.31$ & $4.50$ \\
$3$ & $0.39$ & $12.51$ & $4.58$ \\
\hline
\end{tabular}
\vspace{-1pt}
\caption{Real-world experimental results with simulated gate visuals.}
\label{tab:dry-run-performance}
\vspace{-2pt}
\end{table}

To evaluate our algorithm in a repeatable way, we first simulate the gate visuals using position estimates obtained via GPS (not used for control). 
We simulate a point cloud for the gate, project it into the camera frame, and use its bounding box to calculate the LOS.
This setup allows us to accurately compute the distance to the gate center as the relative position of the quadrotor and the gate are known.
We use~${k_\text{PN}=2}$ and~${\bar{\gamma}=15^\circ}$ and conduct three runs with the linear slow gate motion, each time starting from a stable hovering state.
During the experiments, a wind speed of about~$3\,\si{m}/\si{s}$ is measured.
All runs are successful, and the results provided in~\cref{tab:dry-run-performance} demonstrate consistent performance at high speeds of more than~$12\,\si{m}/\si{s}$.

%
\subsubsection{End-to-End Experiment}
\input{figures/real_world_trajectories}
Due to the challenges of moving a gate at a desired speed without motion capture, we emulate the gate movement with a truck. 
We use YOLO~\cite{redmon2016you} for detection and Nanotrack~\cite{chen2020siamese, yan2021lighttrack} for tracking and use~${k_\text{PN}=2}$ and~$\bar{\gamma}=20^\circ$. We smooth the tracking estimates to prevent LOS jumps that could destabilize the algorithm.
The perception pipeline has a latency of about~$60\,\si{ms}$.
The truck drives at a speed of $8$~to~$10\,\si{m}/\si{s}$, and the quadrotor is initially~$40\,\si{m}$ away.
Once the truck is successfully detected and tracked, we start our drone racing algorithm.
We pull up the quadrotor right before impact to avoid damage.
By extrapolating the quadrotor trajectory based on the last velocity before pulling up, we can predict a distance of~$0.9\,\si{m}$ to the center of the truck's bounding box; see~\cref{fig:real_world_trajectories}. 
The duration of the approach, including the extrapolation, is~$4.9\,\si{s}$, and the quadrotor reaches a speed of~$23.6\,\si{m}/\si{s}$, which is only slightly below the speed achieved in~\cite{kaufmann_champion-level_2023} under controlled conditions. 
The experiment demonstrates that our control algorithm can be integrated with a computer vision stack on real hardware and achieve high speeds while being robust to sensor noise, delays in perception and control, wind, and unmodeled aerodynamic effects.

\section{Conclusion and Future Work}
We have proposed a control algorithm for drone racing with moving gates based solely on monocular LOS and IMU measurements.
We demonstrate through simulation and real-world experiments that our approach is effective at flying through moving gates at high speeds without knowing their relative position and velocity and robust to different gate motions, noise, wind, and delays.
While we have shown that keeping the gate within the field of view can be enforced with our controller, the necessity for this condition is still a limitation of our method.
Our experiments are focused on drone racing, but we consider this work an important step toward vision-based agile flight in dynamic environments without relying on accurate maps and full state estimation.

%

\bibliographystyle{IEEEtran}
\bibliography{ref.bib}

\end{document}